\documentclass[table,dvipsnames,svgnames]{article}  

\usepackage{aaai25}
\usepackage{times}  
\usepackage{helvet}  
\usepackage{courier}  
\usepackage[hyphens]{url}  
\usepackage{graphicx} 
\usepackage{natbib}  
\usepackage{caption} 
\usepackage{algorithm}
\usepackage{xspace} 
\usepackage{graphicx}
\usepackage{multirow}
\usepackage{booktabs}
\usepackage{adjustbox} 
\usepackage{caption}
\usepackage{pdflscape} 
\usepackage[table]{xcolor}
\usepackage{verbatim}
\usepackage{mathptmx} 
\usepackage{amsmath,amssymb,amsfonts}
\usepackage[noend]{algpseudocode}
\usepackage{xcolor, soul}
\sethlcolor{yellow}
\usepackage{siunitx}
\usepackage{subcaption}
\usepackage{mathrsfs}
\usepackage{makecell} 

\usepackage{xcolor}

\PassOptionsToPackage{dvipsnames,svgnames,table}{xcolor}
\usepackage{natbib}
\usepackage{cleveref}

\usepackage{newfloat}
\usepackage{listings}
\DeclareCaptionStyle{ruled}{labelfont=normalfont,labelsep=colon,strut=off} 
\floatstyle{ruled}
\newfloat{listing}{tb}{lst}{}
\floatname{listing}{Listing}
\title{GPSMamba: A Global Phase and Spectral Prompt-guided Mamba for Infrared Image Super-Resolution}
\author {
    Yongsong Huang\textsuperscript{\rm 1},
    Tomo Miyazaki\textsuperscript{\rm 1},
    Xiaofeng Liu\textsuperscript{\rm 2},
    Shinichiro Omachi\textsuperscript{\rm 1}
}
\affiliations {
    \textsuperscript{\rm 1}Tohoku University, Sendai 9808579, Japan\\
    \textsuperscript{\rm 2}Yale University, New Haven 06519, USA\\
    (hys, tomo, shinichiro.omachi.b5)@tohoku.ac.jp,
    xiaofeng.liu@yale.edu
}

\usepackage{bibentry}

\begin{document}

\maketitle

\begin{abstract}
Infrared Image Super-Resolution (IRSR) is challenged by the low contrast and sparse textures of infrared data, requiring robust long-range modeling to maintain global coherence. While State-Space Models like Mamba offer proficiency in modeling long-range dependencies for this task, their inherent 1D causal scanning mechanism fragments the global context of 2D images, hindering fine-detail restoration. To address this, we propose Global Phase and Spectral Prompt-guided Mamba (GPSMamba), a framework that synergizes architectural guidance with non-causal supervision. First, our Adaptive Semantic-Frequency State Space Module (ASF-SSM) injects a fused semantic-frequency prompt directly into the Mamba block, integrating non-local context to guide reconstruction. Then, a novel Thermal-Spectral Attention and Phase Consistency Loss provides explicit, non-causal supervision to enforce global structural and spectral fidelity. By combining these two innovations, our work presents a systematic strategy to mitigate the limitations of causal modeling. Extensive experiments demonstrate that GPSMamba achieves state-of-the-art performance, validating our approach as a powerful new paradigm for infrared image restoration. Code is available at https://github.com/yongsongH/GPSMamba.
\end{abstract}

%

\section{Introduction}
\label{sec:intro}

\begin{figure}[t]
\centerline{\includegraphics[width=\columnwidth]{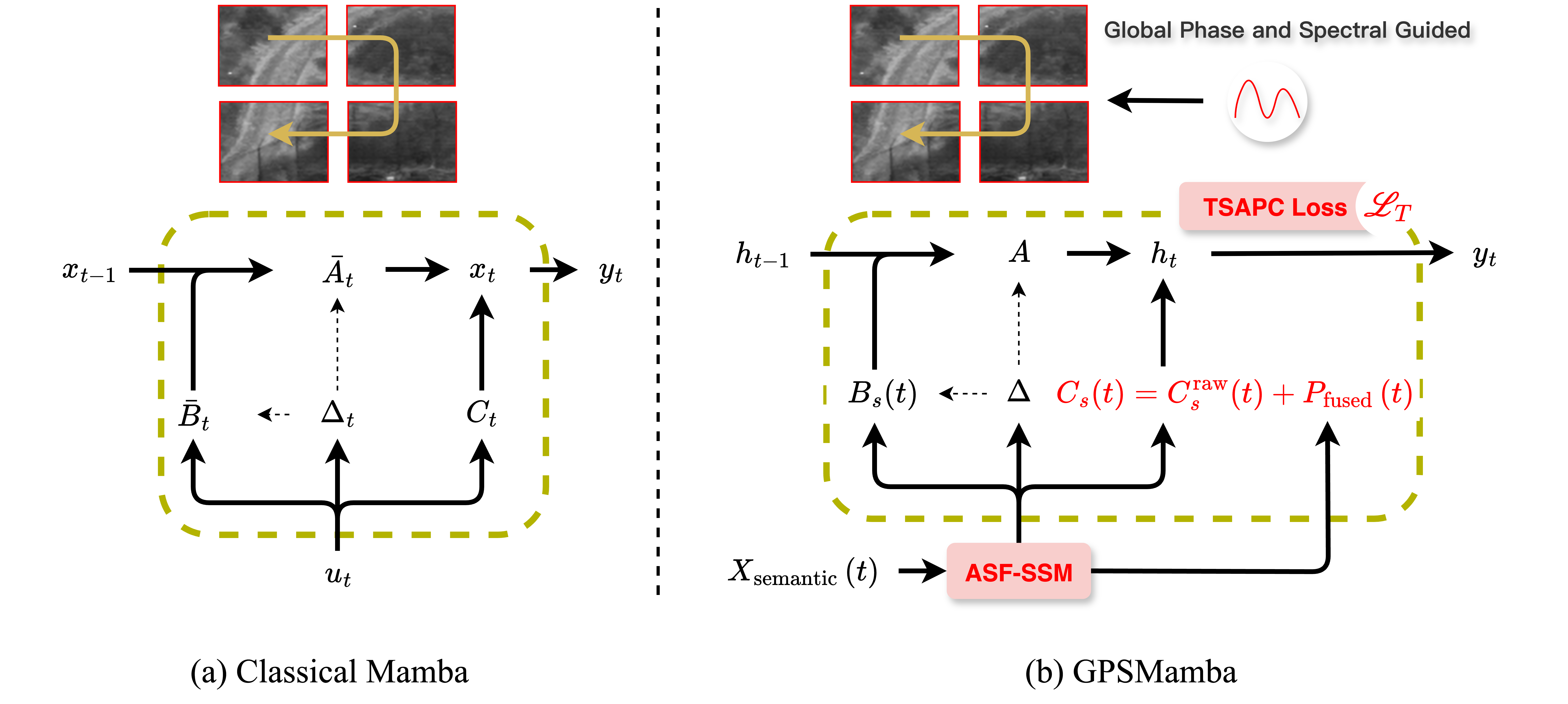}}
\caption{An overview of our proposed GPSMamba for IRSR. (a) Classical Mamba\cite{xiao2025spatialmamba}. The fixed, causal scan path of classical Mamba struggles to capture long-range dependencies in low-contrast infrared images, limiting reconstruction fidelity. (b) GPSMamba. To overcome this, GPSMamba introduces two key innovations: 1) The Adaptive Semantic Fusion State Space Model (ASF-SSM) injects a global frequency-domain prompt ($P_{\text{fused}}$) via an adaptive scan path, breaking the rigid causal chain to integrate non-local context. 2) A novel Thermal-Spectral Attention and Phase Consistency (TSAPC) Loss ($\mathcal{L}_T$) provides explicit, non-causal supervision to enforce global structural and spectral fidelity.}
\label{Fig.1 Overview}
\end{figure}

Infrared (IR) imaging is indispensable in applications such as security surveillance, remote sensing\cite{zhu2022pkulast}, and autonomous driving\cite{tan2024driver}, particularly in environments where visible-light imaging is compromised by poor illumination or adverse weather conditions\cite{fang2024online}. However, raw IR images frequently suffer from low spatial resolution, noise contamination, and poor contrast, stemming from sensor hardware limitations and atmospheric interference\cite{huang2023infrared}. The objective of IRSR is to reconstruct a high-resolution (HR) image, $I_{HR} \in \mathbb{R}^{H \times W \times C}$, from its low-resolution (LR) counterpart, $I_{LR} \in \mathbb{R}^{h \times w \times C}$, where $H, W, C$ denote the height, width, and channel count, and $s$ is the scaling factor ($h=H/s, w=W/s$). This process can be formulated as learning a mapping function $F$ such that $I_{SR} = F(I_{LR})$, where $I_{SR}$ is the reconstructed image that aims to approximate the $I_{HR}$.

Unlike natural image super-resolution, IRSR presents unique challenges due to the intrinsic properties of thermal radiation\cite{11059944}. IR images are often characterized by low contrast, homogeneous backgrounds, and sparse high-frequency details, making it difficult for conventional models to restore fine textures. Furthermore, targets of interest may be spatially distant from each other, demanding that the reconstruction model possesses a superior capability for long-range contextual awareness to ensure global structural coherence\cite{li2025difiisr}. With the advent of deep learning, data-driven methods have become the dominant paradigm in super-resolution\cite{liu2022blind}. Their primary advantage lies in the ability to learn the complex, non-linear mapping from LR to HR feature spaces directly from data, obviating the need for meticulously designed, hand-crafted priors\cite{wang2020deep}. This has established deep learning as the mainstream approach for image restoration tasks. Recently, State-Space Models (SSMs)\cite{guo2024mambair, zhu2024vision}, particularly Mamba\cite{gu2023mamba}, have emerged as a powerful architecture for sequence modeling. Their ability to capture long-range dependencies with linear computational complexity presents a compelling alternative to high-complexity Transformers and GANs, which can be prone to training instability. The potential of SSMs has been successfully demonstrated in image restoration. For instance, MambaIR\cite{guo2024mambair} first validated the effectiveness of Mamba in this domain. MambaIRv2\cite{guo2025mambairv2} further enhanced global context modeling by using similarity maps to guide the model's attention toward semantically relevant regions. Most recently, IRSRMamba\cite{11059944} achieved state-of-the-art (SOTA) performance on several IRSR benchmarks by integrating global priors from the frequency domain.

Despite this progress, we observe that the intrinsic causal modeling of SSMs still poses a fundamental challenge for IRSR. The fixed, 1D causal scanning mechanism of Mamba inherently flattens a 2D image, fragmenting its global structural information into discrete and non-adjacent patches\cite{guo2025mambairv2,xiao2025spatialmamba,shao2025hybrid}. To faithfully restore fine-grained details and maintain global coherence in IR images, it is crucial to develop an SSM-based method that can better aggregate information from semantically relevant regions across the entire image (see Fig.\ref{Fig.1 Overview}). In this work, we propose GPSMamba to address this limitation. As illustrated in Fig. \ref{Fig.1 Overview}, our approach introduces two core innovations: 1) a Mamba architecture enhanced by a global frequency-domain prompt to guide the reconstruction with non-local context, and 2) a novel non-causal loss function designed to explicitly enforce global structural and spectral fidelity. More details will be shown in the method section. Our contributions are as follows:
\begin{itemize}
    \item An Adaptive Semantic-Frequency State Space Module that injects a global frequency-domain prompt to guide the Mamba block, breaking its causal modeling limitations to enhance non-local context fusion.
    \item We also introduce a novel Thermal-Spectral Attention and Phase Consistency Loss ($\mathcal{L}_T$), which provides explicit, non-causal supervision. This loss directly compensates for the information fragmentation caused by causal scanning by enforcing global structural and spectral fidelity across the image.
    \item  GPSMamba achieves new SOTA performance on benchmark IRSR datasets, establishing a more effective paradigm for Mamba-based infrared image restoration.
    
\end{itemize}

\section{Related Work}
\label{sec:rw}

\begin{figure*}[t] 
    \centering
    \includegraphics[width=\linewidth]{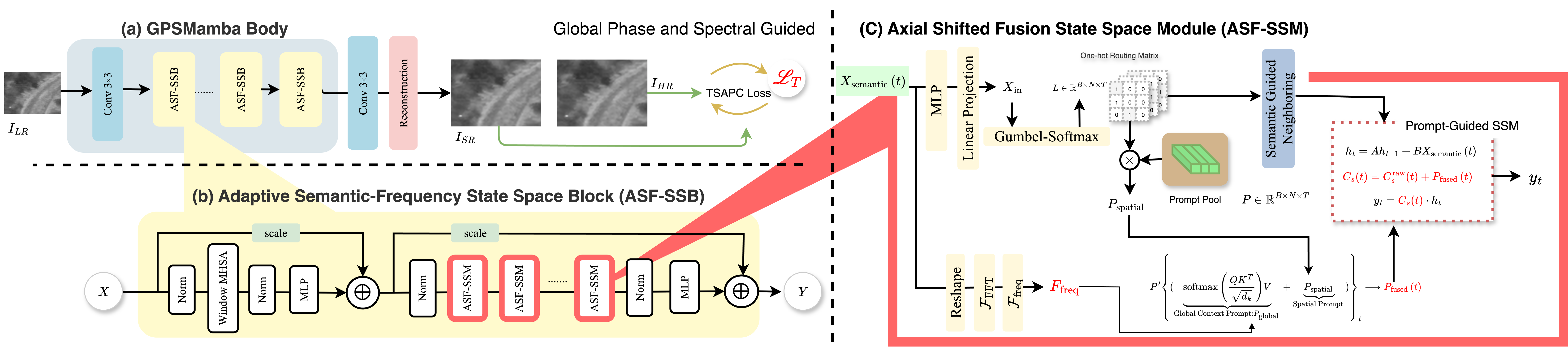}

    \caption{The overall architecture of our proposed GPSMamba model. \textbf{(a) GPSMamba Body:} The top-level architecture processes $I_{LR}$ through a series of Adaptive Semantic-Frequency State Space Blocks (ASF-SSB) to produce the $I_{SR}$. The entire network is trained end-to-end under the supervision of our proposed TSAPC Loss ($\mathcal{L}_T$).  \textbf{(b) Adaptive Semantic-Frequency State Space Block (ASF-SSB):} This is the main building block of the network. It takes an input feature map $X$ and refines it to produce an output $Y$ via a residual connection. Each ASF-SSB contains several of our core ASF-SSM modules, which are responsible for the SSMs scan mechanism.  \textbf{(c) Axial Shifted Fusion State Space Module (ASF-SSM):} It first dynamically generates a complete prompt tensor $P'$ by fusing two components: (1) a content-aware spatial prompt, $P_{\text{spatial}}$, selected from a learnable prompt pool $P$ using a Gumbel-Softmax router, and (2) a global context prompt $P_{\text {global}}$ derived from frequency-domain features $F_{\text{freq}}$ via self-attention. Then, at each timestep $t$, a time-specific vector is selected, $P_{\text{fused}}(t) = P'_{t}$, and injected to guide the state transition. This fusion allows the module to model both local semantics and global structure simultaneously.}
    \label{fig:Overall} 
\end{figure*}

IRSR is an inherently challenging task due to the unique properties of IR data, such as low contrast, high noise levels, and large homogeneous regions with sparse textural details. Deep learning approaches have primarily followed two main architectural paradigms to address this. While Convolutional Neural Network (CNN) based methods are efficient at extracting local features, their limited receptive fields hinder the modeling of long-range dependencies, which is crucial for maintaining structural coherence in IR images\cite{shi2024swinibsr,qin2024lkformer,wu2023swinipisr}. To overcome this limitation, Transformer-based architectures were introduced, leveraging self-attention mechanisms to effectively capture global context and achieve state-of-the-art performance. However, the quadratic computational complexity of self-attention makes these models impractical for HR applications and resource-constrained scenarios\cite{qin2024lkformer}. This establishes a critical trade-off between global modeling capability and computational efficiency. Therefore, there is a pressing need for a new architecture that can efficiently model long-range dependencies for high-fidelity IRSR, motivating our exploration of SSMs.

\subsection{State-Space Models for Image Restoration}
SSMs, particularly the recently proposed Mamba, have emerged as a compelling alternative, offering linear complexity while maintaining the ability to model long-range dependencies. Mamba is built upon a structured SSMs that map a 1D input sequence $x_{t}$ to an output $y(t)$ via a latent state $h_{t}$, governed by the state equations:
\begin{equation}
    h_{t} = \mathbf{A}h_{t-1} + \mathbf{B}x_{t}, \quad y_{t} = \mathbf{C}h_{t}
\end{equation}
where its core innovation, the Selective Scan Mechanism (S6), makes the state matrices $\mathbf{A}, \mathbf{B}, \mathbf{C}$ input-dependent\cite{zhang2024survey, wang2024state}. This allows the model to dynamically focus on relevant information within the sequence. 
The efficiency and power of Mamba have led to its rapid adoption in computer vision. For image restoration, models like MambaIR\cite{guo2024mambair}, MambaIRv2\cite{guo2025mambairv2}, and IRSRMamba\cite{11059944} have successfully applied SSMs, achieving SOTA results. These methods typically adapt Mamba to 2D data by flattening image patches into 1D sequences and processing them with bidirectional scans. However, this mandatory flattening and the inherent causal nature of the scanning mechanism can disrupt the spatial contiguity of the 2D image structure\cite{xiao2025spatialmamba,shao2025hybrid}. This fragmentation of global context is a significant drawback, especially for IRSR, where restoring sparse textures relies heavily on a holistic understanding of the entire scene. While existing works have attempted to mitigate this with multi-scale feature fusion or semantic guidance, these are often post-hoc corrections. Our work, GPSMamba, addresses this fundamental issue by directly integrating a global prompt into the SSM's scanning process, enabling non-local information to guide feature extraction from the outset.
\subsection{Global Priors and Guidance for Restoration}
Incorporating priors is a well-established strategy to regularize the ill-posed nature of image restoration. Frequency-domain information, for example, serves as a powerful global prior. The Fourier spectrum decouples an image's global structure (magnitude) from its fine-grained details (phase), which has been exploited by various methods to improve reconstruction fidelity\cite{xiao2024bridging,xiong2024rshazediff}. Other forms of guidance have also been explored. For instance, some works have designed uncertainty-driven loss functions to make the training process focus on more challenging regions\cite{peng2024uncertainty}. More recently, techniques like all-in-one restoration models have been developed by adapting large pre-trained networks with minimal overhead\cite{qin2024restore,yao2024neural}. Our work introduces a novel form of guidance by framing the global prior as a dynamic prompt. Unlike methods that use frequency information as a regularization loss or in a separate processing branch, we inject a frequency-domain prompt directly into the core of our proposed ASF-SSM. This prompt actively modulates the state transitions within the Mamba block, making the feature extraction process aware of the global image context at every step. This tight integration of prompt-based guidance with the SSM architecture is a key distinction from previous works. 

\section{Method}
\label{sec:method}

Our proposed method, GPSMamba, is a hierarchical network designed for IRSR, with its overall architecture illustrated in Fig. \ref{fig:Overall}. The network first employs a shallow convolutional layer to extract low-level features from the $I_{LR}$ input. The core of the model is a deep feature extraction body composed of several stacked ASF-SSBs. The central innovation lies within our ASF-SSM, which enhances the standard state-space mechanism with a dynamic, content-aware prompt generated from both semantic and spectral information. Finally, a reconstruction module upscales the learned deep features to produce the $I_{SR}$. To guide the network's training, we introduce a novel objective function, the Thermal-Spectral Attention and Phase Consistency (TSAPC) Loss ($\mathcal{L}_T$). The following subsections provide a detailed exposition of our proposed ASF-SSM and the formulation of the TSAPC Loss.

\begin{algorithm}[h]
\caption{ASF-SSM Core Processing}
\label{alg:ASF-SSM_concise}
\begin{algorithmic}[1]
\Require{ $X_{\text{semantic}} \in \mathbb{R}^{B \times N \times C}$,  $P \in \mathbb{R}^{B \times N \times T}$, $Y_{\text{scan}}$}
\Ensure{ $Y \in \mathbb{R}^{B \times N \times C}$}

\State $X_\text{in} \gets \text{Linear Projection}(\text{MLP}(X_{\text{semantic}}))$
\State $L \gets \text{Gumbel-Softmax}(X_\text{in}, \text{hard=True})$
\State $P_{\text{spatial}} \gets L \cdot P$
\State $F_{\text{freq}} \gets F_{\mathrm{FFT}}(\text{Reshape}(X_{\text{semantic}}))$
\State $Q, K, V \gets \text{Linear}_{QKV}(F_{\text{freq}})$
\State $P_{\text{global}} \gets \text{softmax}\left(\frac{QK^T}{\sqrt{d_k}}\right) V$
\State $P' \gets P_{\text{spatial}} + P_{\text{global}}$

\State $\Delta, B, C_{\text{raw}} \gets \text{split}(X_{\text{in}})$
\State $A \gets -\exp(\text{Linear}_{\Delta}(\Delta))$

\State $h_0 \gets \mathbf{0} \in \mathbb{R}^{B \times C}$

\For{$t=1 \ldots N$}
    \State $P_{\text{fused}}(t) \gets P'(t)$
    \State $C_{s}(t) \gets C_{\text{raw}}(t) + P_{\text{fused}}(t)$
    \State $h_t \gets A(t) \odot h_{t-1} + B(t) \odot X_{\text{semantic}}(t)$
    \State $y_t \gets C_{s} \odot h_t$
    \State Append $y_t$ to $Y_{\text{scan}}$
\EndFor

\State $Y \gets \text{Linear}_{\text{out}}(\text{LayerNorm}(Y_{\text{scan}}))$
\State \Return $Y$
\end{algorithmic}
\end{algorithm}

\subsection{Adaptive Semantic-Frequency State Space Module}

The Adaptive Semantic-Frequency State Space Module (ASF-SSM) is the cornerstone of our architecture, designed to overcome a key limitation of standard SSMs. While models like Mamba achieve linear complexity by processing flattened image patches as 1D sequences, this serialization can disrupt the 2D spatial relationships and global frequency patterns crucial for image understanding. Our ASF-SSM addresses this by enriching the selective scan mechanism with a dynamic, multi-domain prompt that guides state transitions based on both local semantics and global context. The ASF-SSM, detailed in Algorithm \ref{alg:ASF-SSM_concise}, comprises two synergistic stages: (a) Fused Prompt Generation, and (b) Prompt-Guided State Space Model.

\textbf{Fused Prompt Generation.} We generate a comprehensive prompt tensor $P'$ by fusing local and global context. This process begins with the input feature sequence $X_{\text{semantic}} \in \mathbb{R}^{B \times N \times C}$, where $B, N,$ and $C$ are the batch, sequence length, and channel dimensions, respectively. To obtain the local component, a content-aware spatial prompt $P_{\text{spatial}}$, we first employ an MLP to predict routing logits from $X_{\text{semantic}}$. These logits are converted into a discrete, one-hot routing matrix $L \in \mathbb{R}^{B \times N \times T}$ via the Gumbel-Softmax function, where $T$ is the size of a learnable prompt pool $P$. The spatial prompt is then computed as $P_{\text{spatial}} = L\cdot P$. Concurrently, to integrate global context, the input $X_{\text{semantic}}$ is first reshaped and then transformed into the frequency domain via a Fast Fourier Transform (FFT), yielding frequency features $F_{\text{freq}}$. Inspired by the ability of the frequency domain to capture holistic image properties, we process these features through an attention mechanism. Specifically, $F_{\text{freq}}$ is linearly projected into query ($Q$), key ($K$), and value ($V$) tensors to compute a context prompt, $P_{\text{global}}$.
These two prompts, computed in parallel, are then fused via element-wise addition to create the final, comprehensive prompt tensor: $P' \gets P_{\text{spatial}} + P_{\text{global}}$.

\textbf{Prompt-Guided State Space Model.} Our core innovation is to guide the SSM's selective scan by dynamically modulating its output projection matrix $C_{s}(t)$. We first compute a comprehensive prompt tensor $P'$ by fusing spatial and global context prompts, $P' = P_{\text{spatial}} + P_{\text{global}}$.

To break the rigid causal chain, we introduce a semantic-guided scan. During this scan, the time-specific vector $P_{\text{fused}}(t)$ is selected from the tensor $P'$ according to this semantic order, which is determined by sorting tokens based on the argmax of the routing matrix $L$ \cite{guo2025mambairv2}. This prompt is then injected into the output projection:
\begin{equation}
C_s(t) = C_s^{\mathrm{raw}}(t) + P_{\text{fused}}(t)
\end{equation}
The other dynamic SSM parameters ($\Delta$, $B(t)$, $C_s^{\mathrm{raw}}(t)$) are derived from a linear projection of the input sequence, $X_{\text{in}} = \text{Linear}(X_{\text{semantic}})$, followed by a split operation. The state update and output generation then follow the standard discretized SSM formulation, with the input sequence $X_{\text{semantic}}$ along the same semantic path:
\begin{align}
h_t &= A(t) \odot h_{t-1} + B(t) \odot X_{\text{semantic}}(t) \\
y_t &= C_{s}(t) \odot h_t
\end{align}
The resulting output sequence $Y_{\text{scan}} = [y_1, \ldots, y_N]$ is then passed through a final LayerNorm and linear projection, before being reordered back to its original spatial arrangement to produce the module's output $Y$.

\begin{table*}[ht]
\centering
\renewcommand\arraystretch{1.1}
\caption{The average results of (PSNR$\uparrow$ MSE$\downarrow$ SSIM$\uparrow$) with scale factor of  4 \& 2 on datasets result-A \& result-C \& CVC10. Best and second-best performances are marked in \textbf{bold} and {\ul underlined}, respectively.}
\resizebox{\textwidth}{!}{%
\begin{tabular}{@{}c|c|c|cccccccccc@{}} 
\toprule
\multirow{2}{*}{Scale}       & \multicolumn{1}{c|}{\multirow{2}{*}{Methods}} & \multicolumn{1}{c|}{\multirow{2}{*}{\# Params. (K)}} & \multicolumn{3}{c|}{result-A}                                              & \multicolumn{3}{c|}{result-C}                                              & \multicolumn{3}{c}{CVC10}                             \\ \cmidrule(l){4-12} 
                             & \multicolumn{1}{c|}{}                         & \multicolumn{1}{c|}{}                         & PSNR$\uparrow $  & MSE$\downarrow $ & \multicolumn{1}{c|}{SSIM$\uparrow $} & PSNR$\uparrow $  & MSE$\downarrow $ & \multicolumn{1}{c|}{SSIM$\uparrow $} & PSNR$\uparrow $  & MSE$\downarrow $ & SSIM$\uparrow $ \\ \midrule
\multirow{19}{*}{$\times 2$} 
                             & EDSR\textcolor[RGB]{217,205,144}{\textit{[CVPRW 2017]}}~\cite{lim2017enhanced}                                            & 1,369                                              & 39.0493          & 11.8196          & 0.9414                               & 39.8902          & 8.9865           & 0.9528                               & 44.1770          & 2.7845           & 0.9713          \\
                             & ESRGAN\textcolor[RGB]{217,205,144}{\textit{[ECCVW 2018]}}~\cite{wang2018esrgan}                                        & 16,661                                               & 38.7738          & 12.5212          & 0.9384                               & 39.6111          & 9.5793           & 0.9500                               & 44.0974          & 2.8477           & 0.9709          \\
                             & FSRCNN\textcolor[RGB]{217,205,144}{\textit{[ECCV 2016]}}~\cite{dong2016accelerating}                                        & 475                                               & 39.1175          & 11.3761          & 0.9426                               & 39.9858          & 8.6899           & 0.9535                               & 44.1253          & 2.8162           & 0.9710          \\
                             & SRGAN\textcolor[RGB]{217,205,144}{\textit{[CVPR 2017]}}~\cite{ledig2017photo}                                        & 1,370                                               & 39.0401          & 11.9024          & 0.9414                               & 39.8678          & 9.0586           & 0.9527                               & 44.1736          & 2.7851           & 0.9713          \\
                             & SwinIR\textcolor[RGB]{217,205,144}{\textit{[ICCV 2021]}}~\cite{liang2021swinir}                                        & 11,752                                               & 38.6899          & 12.5694          & 0.9374                               & 39.5215          & 9.6530           & 0.9492                               & 43.9980          & 2.8926           & 0.9704          \\
                             & SRCNN\textcolor[RGB]{217,205,144}{\textit{[T-PAMI 2015]}}~\cite{dong2015image}                                          & 57                                               & 38.9671          & 11.7216          & 0.9414                               & 39.8642          & 8.8857           & 0.9524                               & 44.0038          & 2.9084           & 0.9707          \\
                             & RCAN\textcolor[RGB]{217,205,144}{\textit{[ECCV 2018]}}~\cite{zhang2018image}                                          & 12,467                                               & 38.8145          & 12.4926          & 0.9391                               & 39.7075          & 9.4220           & 0.9511                               & 44.1205          & 2.8170           & 0.9713          \\
                             & PSRGAN\textcolor[RGB]{217,205,144}{\textit{[SPL 2021]}}~\cite{huang2021infrared}                                       & 2,414                                               & 39.2146          & 11.2409          & 0.9429                               & 40.0543          & 8.6101           & 0.9539                               & 44.2377          & 2.7454           & 0.9713          \\
                             & ShuffleMixer(tiny)\textcolor[RGB]{217,205,144}{\textit{[NIPS'22]}}~\cite{sun2022shufflemixer})                            & 108                                               & 39.0465          & 11.7605          & 0.9414                               & 39.8766          & 8.9680           & 0.9527                               & 44.1408          & 2.8113           & 0.9713          \\
                             & ShuffleMixer (base)\textcolor[RGB]{217,205,144}{\textit{[NIPS'22]}}~\cite{sun2022shufflemixer}                            & 121                                               & 38.8066          & 12.3718          & 0.9388                               & 39.6347          & 9.4864           & 0.9503                               & 44.0357          & 2.8809           & 0.9710          \\
                             & HAT\textcolor[RGB]{217,205,144}{\textit{[CVPR 2023]}}~\cite{Chen_2023_CVPR}                                           & 20,624                                               & 38.7754          & 12.4528          & 0.9384                               & 39.6346          & 9.5132           & 0.9500                               & 44.1080          & 2.8244           & 0.9709          \\
                             & RGT\textcolor[RGB]{217,205,144}{\textit{[ICLR 2024]}}~\cite{chen2024recursive}                                              & 10,051                                               & 39.1642          & 11.3382          & 0.9429                               & 40.0522          & 8.6033           & 0.9540                               & 44.2311          & 2.7358           & 0.9717          \\
                             & MambaIR \textcolor[RGB]{217,205,144}{\textit{[ECCV 2024 SOTA]}}~\cite{guo2024mambair}                                              & 20,421                                               & 39.1761          & 11.2081          & \underline{0.9437}                               & 40.1399          & 8.4798           & \underline{0.9544}                              & 44.4181          & 2.6076           & \textbf{0.9720}          \\
                             & ATD\textcolor[RGB]{217,205,144}{\textit{[CVPR 2024]}}~\cite{zhang2024transcending}                                           & 753                                               & 39.0453          & 11.5702          & 0.9432                               & 40.0375          & 8.6155           & 0.9542                              & 44.1901          & 2.7737           & 0.9711          \\
                             & CATANet \textcolor[RGB]{217,205,144}{\textit{[CVPR 2025 SOTA]}}~\cite{liu2025catanet}                                            & 477                                               & 39.0886 & 11.4332 & 0.9430                    & 40.0064 & 8.6387  & 0.9539                      & 44.1656 & 2.7858  & 0.9712 \\
                             & MambaOut\textcolor[RGB]{217,205,144}{\textit{[CVPR 2025 SOTA]}}\cite{yu2024mambaout}                 & 9,669                        & 38.6375          & 12.7091          & 0.9371                               & 39.4900          & 9.7035           & 0.9493                               & 43.9150          & 2.9429           & 0.9704          \\
                             & VisionMamba\textcolor[RGB]{217,205,144}{\textit{[ICML 2024]}}\cite{zhu2024vision}                   &  27,880                        & 38.7805          & 12.2990          & 0.9392                               & 39.6339          & 9.3781           & 0.9506                               & 43.9521          & 2.9103           & 0.9704          \\
                             & IRSRMamba\textcolor[RGB]{217,205,144}{\textit{[TGRS 2025 SOTA]}}~\cite{11059944}                                           & 26,462                                               & \underline{39.3489} & \underline{10.8767} & \textbf{0.9440}                      & \underline{40.2302} & \underline{8.3164}  & \textbf{0.9548}                      & \underline{44.5310} & \underline{2.5537}  & \textbf{0.9720} \\
                             &  MambaIRv2 \textcolor[RGB]{217,205,144}{\textit{[CVPR 2025 SOTA]}}~\cite{guo2025mambairv2}                                                                       & 22,903                                               &  39.0421                & 11.5690                  &  0.9423                                    & 39.9295                & 8.7817                 & 0.9531                                    & 44.1442                 & 2.7995                 & 0.9711                 \\
                             & \textbf{GPSMamba (Ours)}                                           & 36,942                                               & \textbf{39.3505} & \textbf{10.8608} & \textbf{0.9440}                      & \textbf{40.2418} & \textbf{8.2943}  & \textbf{0.9548}                      & \textbf{44.5543} & \textbf{2.5451}  & \underline{0.9719} \\

                             \midrule
\multirow{19}{*}{$\times 4$} 
                             & EDSR\textcolor[RGB]{217,205,144}{\textit{[CVPRW 2017]}}~\cite{lim2017enhanced}                                            & 1,369                                              & 34.5219          & 30.1273          & 0.8548                               & 35.1740          & 23.9917          & 0.8723                              & 40.1190          & 6.8819           & 0.9482          \\
                             & ESRGAN\textcolor[RGB]{217,205,144}{\textit{[ECCVW 2018]}}~\cite{wang2018esrgan}                                         & 16,661                                               & 33.6895          & 34.7337          & 0.8500                               & 34.1650          & 28.9017          & 0.8679                               & 37.9780          & 10.9641          & 0.9455          \\
                             & FSRCNN\textcolor[RGB]{217,205,144}{\textit{[ECCV 2016]}}~\cite{dong2016accelerating}                                        & 475                                               & 33.8556          & 34.4909          & 0.8446                              & 34.5272          & 27.4495          & 0.8636                               & 38.7856          & 9.5482           & 0.9421          \\
                             & SRGAN\textcolor[RGB]{217,205,144}{\textit{[CVPR 2017]}}~\cite{ledig2017photo}                                        & 1,370                                               & 34.5807          & 29.6927         & 0.8556                               & 35.2076          & 23.7701          & 0.8728                               & 40.1479          & 6.8162           & 0.9483          \\
                             & SwinIR\textcolor[RGB]{217,205,144}{\textit{[ICCV 2021]}}~\cite{liang2021swinir}                                        & 11,752                                               & 34.4321          & 30.6081          & 0.8537                               & 35.0329          & 24.6490          & 0.8710                               & 39.9062          & 7.1886           & 0.9479          \\
                             & SRCNN\textcolor[RGB]{217,205,144}{\textit{[T-PAMI 2015]}}~\cite{dong2015image}                                          & 57                                               & 33.6839          & 34.9181          & 0.8415                               & 34.2348          & 28.6115          & 0.8568                               & 38.0976          & 10.7588          & 0.9279          \\
                             & RCAN\textcolor[RGB]{217,205,144}{\textit{[ECCV 2018]}}~\cite{zhang2018image}                                          & 12,467                                               & 34.4280          & 30.8815          & 0.8528                               & 35.0823          & 24.6507          & 0.8705                               & 40.0805          & 6.9225           & 0.9484          \\
                             & PSRGAN\textcolor[RGB]{217,205,144}{\textit{[SPL 2021]}}~\cite{huang2021infrared}                                        & 2,414                                               & 34.4595          & 30.3760          & 0.8540                               & 35.1023          & 24.3147          & 0.8715                               & 39.9533          & 7.1274           & 0.9471          \\
                             & ShuffleMixer(tiny)\textcolor[RGB]{217,205,144}{\textit{[NIPS'22]}}~\cite{sun2022shufflemixer})                            & 108                                               & 34.5440          & 29.9449          & 0.8550                               & 35.1640          & 23.9705          & 0.8723                               & 40.0756          & 6.9296           & 0.9478          \\
                             & ShuffleMixer (base)\textcolor[RGB]{217,205,144}{\textit{[NIPS'22]}}~\cite{sun2022shufflemixer}                            & 121                                               & 34.4507          & 30.6955          & 0.8538                                & 35.0911          & 24.3745          & 0.8714                               & 40.0120          & 7.0622           & 0.9477          \\
                             & HAT\textcolor[RGB]{217,205,144}{\textit{[CVPR 2023]}}~\cite{Chen_2023_CVPR}                                           & 20,624                                               & 34.4947          & 30.4086          & 0.8542                               & 35.1239          & 24.3103          & 0.8713                               & 40.0934          & 6.9078           & 0.9478          \\
                & RGT\textcolor[RGB]{217,205,144}{\textit{[ICLR 2024]}}~\cite{chen2024recursive}                                             & 10,051                                               & 34.3826          & 31.0046          & 0.8535          & 35.0534          & 24.5924          & 0.8711                & 39.8420          & 7.3060           & 0.9472           \\
                            & MambaIR\textcolor[RGB]{217,205,144}{\textit{[ECCV 2024 SOTA]}}~\cite{guo2024mambair}                                           & 20,421                                               & 34.0267          & 32.9760          & 0.8510                               & 34.5662          & 27.0850          & 0.8681                               & 38.1878          & 10.8653          & 0.9404          \\
                              & ATD\textcolor[RGB]{217,205,144}{\textit{[CVPR 2024]}}~\cite{zhang2024transcending}                                           & 753                                               & 34.6113          & 29.5294          & 0.8569                              & 35.2347          & 23.6882          & 0.8737                              & 40.2897          & 6.6001           & 0.9494      \\
                            & CATANet \textcolor[RGB]{217,205,144}{\textit{[CVPR 2025 SOTA]}}~\cite{liu2025catanet}                                           & 477                                               & 34.0613 & 32.7100  & 0.8447                    & 34.8947 & 25.3673  & 0.8663                      & 39.0302 & 8.8215  & 0.9405 \\
                             & MambaOut\textcolor[RGB]{217,205,144}{\textit{[CVPR 2025 SOTA]}}\cite{yu2024mambaout}                 & 9,669                        & 34.4483          & 30.6792          & 0.8527                               & 35.0456          & 24.7055          & 0.8698                               & 40.1587          & 6.8076           & 0.9485          \\
                             & VisionMamba\textcolor[RGB]{217,205,144}{\textit{[ICML 2024]}}\cite{zhu2024vision}                   & 27,880                        & 34.5941          & 29.5650          & 0.8564                               & 35.2327          & 23.6467          & 0.8733                               & 40.1705          & 6.7933           & 0.9484          \\
                             & IRSRMamba\textcolor[RGB]{217,205,144}{\textit{[TGRS 2025 SOTA]}}~\cite{11059944}                                     & 26,462                                               & \underline{34.6755} & \underline{29.0551} & \underline{0.8577}                      & \underline{35.3074} & \underline{23.2857} & \underline{0.8745}                      & \underline{40.4052} & \underline{6.4536}  & \underline{0.9497} \\
                             &  MambaIRv2 \textcolor[RGB]{217,205,144}{\textit{[CVPR 2025 SOTA]}}~\cite{guo2025mambairv2}                                                                       & 22,903                                               &  29.5295                &  97.3011                &  0.8420                                    &  28.6532                & 130.8167                  & 0.8577                                     & 27.2805                 &  121.7986                &  0.9315               \\
                             & \textbf{GPSMamba (Ours)}                                           & 36,942                                               & \textbf{34.7421} & \textbf{28.7268} & \textbf{0.8587}                      & \textbf{35.4007} & \textbf{22.9636} & \textbf{0.8756}                      & \textbf{40.5475} & \textbf{6.2681}  & \textbf{0.9503} \\

                             \bottomrule
\end{tabular}%
}
\label{tab.2}
\end{table*}

\subsection{TSAPC Loss: $\mathcal{L}_T$}

In IRSR, conventional pixel-wise losses (e.g., L1 or L2) often fail to capture global structural coherence and can struggle to reconstruct faint but critical thermal signatures. This limitation is particularly pronounced for sequential models, such as Mamba, which benefit from global supervisory signals. To address this, we introduce the Thermal-Spectral Attention and Phase Consistency (TSAPC) Loss, a composite objective function designed to enhance both structural fidelity and the representation of salient thermal features in the reconstructed image $I_{SR}$.

Our proposed loss, denoted as $\mathcal{L}_{\text{T}}$, is a weighted sum of two frequency-domain components:
\begin{equation}
    \mathcal{L}_{\text{T}} =  \lambda_{\text{phase}} \mathcal{L}_{\text{phase}} + \lambda_{\text{freq}} \mathcal{L}_{\text{freq}}
\end{equation}
where $\mathcal{L}_{\text{freq}}$ and $\mathcal{L}_{\text{phase}}$ are the two loss components, and $\lambda_{\text{freq}}$ and $\lambda_{\text{phase}}$ are their respective scalar weights. We now detail each component.

The first component, the Phase Consistency Loss ($\mathcal{L}_{\text{phase}}$), operates in the Fourier domain to enforce high-level structural correspondence between the $I_{SR}$ and  $I_{HR}$. It is defined as the L1 distance between their phase angles:
\begin{equation}
\mathcal{L}_{\text{phase}} = \left\| \text{angle}(F_{FFT}(I_{SR})) - \text{angle}(F_{FFT}(I_{HR})) \right\|_1
\end{equation}
where $F_{FFT}(\cdot)$ represents the 2D FFT and $\text{angle}(\cdot)$ extracts the phase component. Since image phase is known to carry critical information about structural layout, this loss term is vital for preserving sharp edges, object boundaries, and overall structural integrity in the reconstructed output.

The second component, the Thermal-Spectral Attention Loss ($\mathcal{L}_{\text{freq}}$), is engineered to focus the model's learning on accurately reconstructing the spectral magnitudes within thermally significant regions. This is achieved by first generating a thermal attention mask, $M_{\text{thermal}}$, from $I_{HR}$. Specifically, $I_{HR}$ is passed through the feature extraction layers of a pre-trained VGG19 network ($\Phi_{\text{VGG}}$), and the resulting features are fed into a lightweight convolutional gating network $G_{\text{gate}}$ with a final sigmoid activation $\sigma(\cdot)$ to produce the mask:
\begin{align}
M_{\text{thermal}} = \sigma\left(G_{\text{gate}}\left(\Phi_{\text{VGG}}(I_{HR})\right)\right)
\end{align}
This mask highlights regions of high thermal activity. After being interpolated to match the spatial dimensions of the images, $M_{\text{thermal}}$ is element-wise multiplied with both $I_{SR}$ and $I_{HR}$. The $\mathcal{L}_{\text{freq}}$ is then computed as the L1 distance between the magnitudes of the FFT of these masked images:
\begin{equation}
\mathcal{L}_{\text{freq}} = \left\| \left|F_{FFT}(I_{SR} \odot M_{\text{thermal}})\right| - \left|F_{FFT}(I_{HR} \odot M_{\text{thermal}})\right| \right\|_1
\end{equation}
where $\odot$ denotes element-wise multiplication and $|\cdot|$ computes the magnitude. This semantically-guided spectral matching compels our model to prioritize the fidelity of frequency components originating from regions identified as thermally salient.

The synergy between $\mathcal{L}_{\text{phase}}$ and $\mathcal{L}_{\text{freq}}$ provides our GPSMamba architecture with both global structural constraints and feature-specific spectral guidance. This dual-component supervision is particularly beneficial for our Mamba-based architecture; while the inherent sequential processing excels at local dependencies, our proposed loss introduces crucial non-local, global supervisory signals. This helps the model transcend the limitations of its causal receptive field, encouraging it to learn representations coherent with the global context of the IR scene. Consequently, the TSAPC-Loss promotes the generation of images with superior detail recovery, enhanced contrast in thermally active regions, and more faithful rendering of complex thermal phenomena.

\section{Experiments}
\label{sec:exp}

In this section, we present a comprehensive evaluation of our proposed GPSMamba. We first detail the experimental setup, including datasets, evaluation metrics, and implementation specifics. We then provide extensive quantitative comparisons against SOTA methods, followed by qualitative visual analysis and in-depth ablation studies to validate our design choices.

\textbf{Datasets.} To ensure a rigorous and fair comparison with SOTA methods, we adopt the experimental protocol established by recent works like IRSRMamba. Our model was trained on the widely-used M3FD dataset\cite{liu2022target}. For evaluation, we benchmark our GPSMamba on three standard infrared test sets: result-A\cite{liu2018infrared, huang2021infrared}, result-C\cite{zhang2017infrared, huang2021infrared}, and CVC10\cite{campo2012multimodal}. Following standard practice, the LR inputs were generated by applying bicubic downsampling to the HR images for scale factors of $\times 2$ and $\times 4$. 

\textbf{Evaluation Metrics.} We assess performance using standard full-reference metrics: Peak Signal-to-Noise Ratio (PSNR/dB) and Structural Similarity Index (SSIM), computed on the Y channel of the YCbCr space. To provide a more holistic assessment of perceptual quality, which is not fully captured by pixel-wise measures, we supplement our analysis with four diverse no-reference (NR-IQA) metrics: the learning-based DBCNN\cite{dbcnn}, the statistical NIQE\cite{niqe} and BRISQUE\cite{BRISQUE}, and the vision-language based CLIP-IQA\cite{clipiqa}. \textbf{Other Settings.} All models were implemented in PyTorch and trained on an NVIDIA A6000 GPU. We optimized the network using the Adam optimizer with a learning rate of $1 \times 10^{-5}$ and a batch size of 32.

\begin{figure*}[hbt!]
    \centering

    \begin{subfigure}[b]{0.85\textwidth} 
        \centering
        \includegraphics[width=\textwidth]{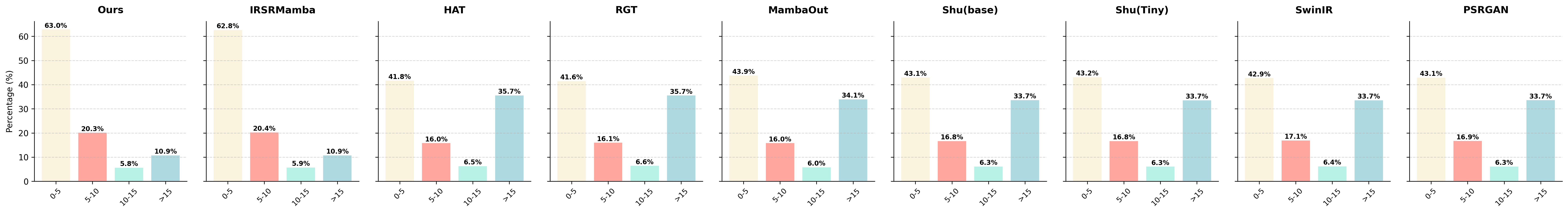} 
        \caption{Error Distribution Across Models on result-A (Scale $\times 4$)} 
        \label{fig:error_distribution_A} 
    \end{subfigure}
    \hfill

    \begin{subfigure}[b]{0.85\textwidth} 
        \centering
        \includegraphics[width=\textwidth]{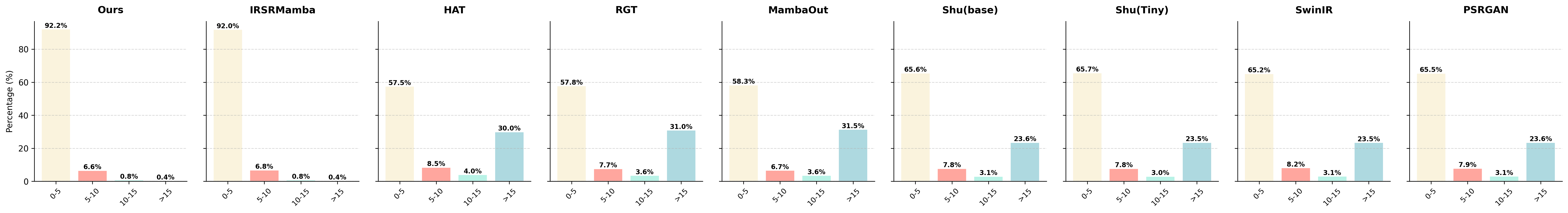} 
        \caption{Error Distribution Across Models on CVC10 (Scale $\times 4$)} 
        \label{fig:error_distribution_C} 
    \end{subfigure}

    \caption{Global error distribution analysis for various super-resolution models on the result-A and CVC10 datasets at scaling factor of $\times 4$. Each subfigure presents a comparative error breakdown across different ranges, quantifying the proportion of minimal, moderate, high, and severe reconstruction errors.}
    \label{fig:error_distribution_combined}
\end{figure*}

\subsection{Experiment Results and Analysis}
\textbf{Quantitative Comparison.} 

\begin{figure}[hbpt]
\centerline{\includegraphics[width=0.8\columnwidth]{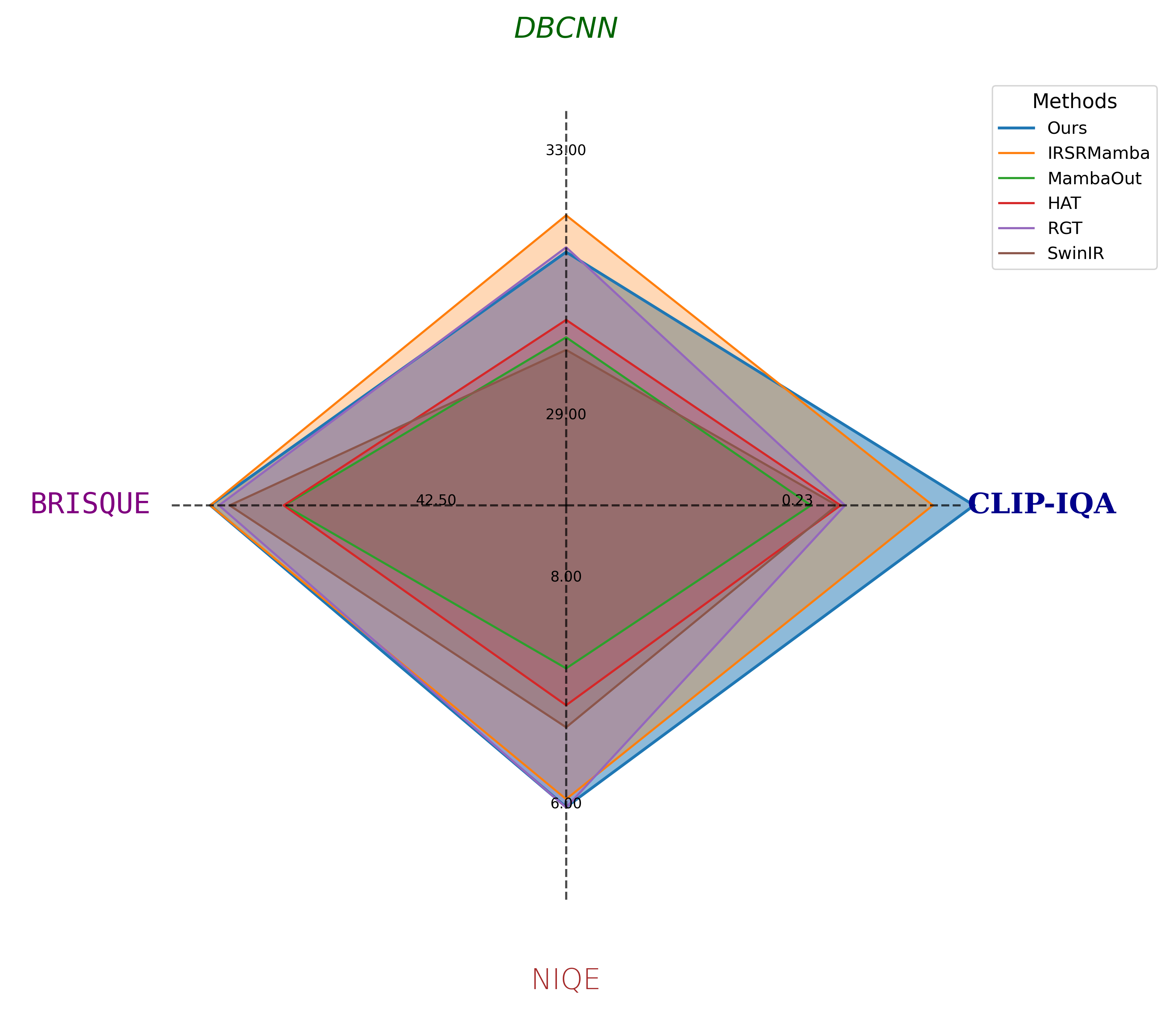}}
\caption{Perceptual quality assessment of super-resolution models on result-A. Larger area indicates better perceptual performance. (Scale $\times 2$)}
\label{PerceptualA}
\end{figure}
We conduct a comprehensive quantitative comparison against a suite of SOTA methods, including recent top-conference models like CATANet, MambaOut, and the current leading architecture, IRSRMamba. The full results for $\times 2$ and $\times 4$ super-resolution on three benchmark datasets are presented in Table \ref{tab.2}. Our proposed model, GPSMamba, demonstrates exceptional performance. For the standard $\times 2$ SR task, it achieves either the best or second-best results across all datasets and metrics, establishing its competitive standing. For instance, on the result-C dataset, our method surpasses IRSRMamba in PSNR and matches its leading SSIM score. The superiority of our approach becomes more pronounced in the more challenging $\times 4$ SR task, where recovering high-frequency details is significantly more difficult. As shown in Table \ref{tab.2}, our GPSMamba consistently sets a new SOTA, outperforming all competing methods on all metrics across all three datasets. On the result-C dataset, we achieve a PSNR of 35.4007 dB, surpassing the previous best (IRSRMamba) by a significant margin of nearly 0.1 dB. This marked improvement underscores our model's enhanced capability to reconstruct faithful and structurally coherent images from highly degraded inputs. To dissect performance beyond aggregate metrics, we analyze the pixel-wise error distribution in Fig. \ref{fig:error_distribution_combined}. The visualization for the $\times 4$ task reveals our model's advantage: it reconstructs 63\% of pixels with minimal error ([0, 5)), surpassing all competitors (result-A (Scale $\times 4$)). Consequently, it produces the fewest pixels in the high-error categories ([10, $\infty$)). To complement our fidelity-based analysis, we assess perceptual quality using four non-reference metrics. As illustrated in Fig. \ref{PerceptualA}, our method's superiority is most evident on the challenging CLIP-IQA metric, where it achieves a leading score of 0.3043 and surpasses the current SOTA, IRSRMamba (0.2887). This advantage, which points to a more semantically coherent and realistic reconstruction, is supported by highly competitive performance across DBCNN, BRISQUE, and NIQE. The radar chart visually synthesizes these findings, showing our model's performance envelope (blue) achieves the most favorable overall balance.

\begin{figure*}[hbt!] 
    \centering
    \includegraphics[width=0.8\linewidth]{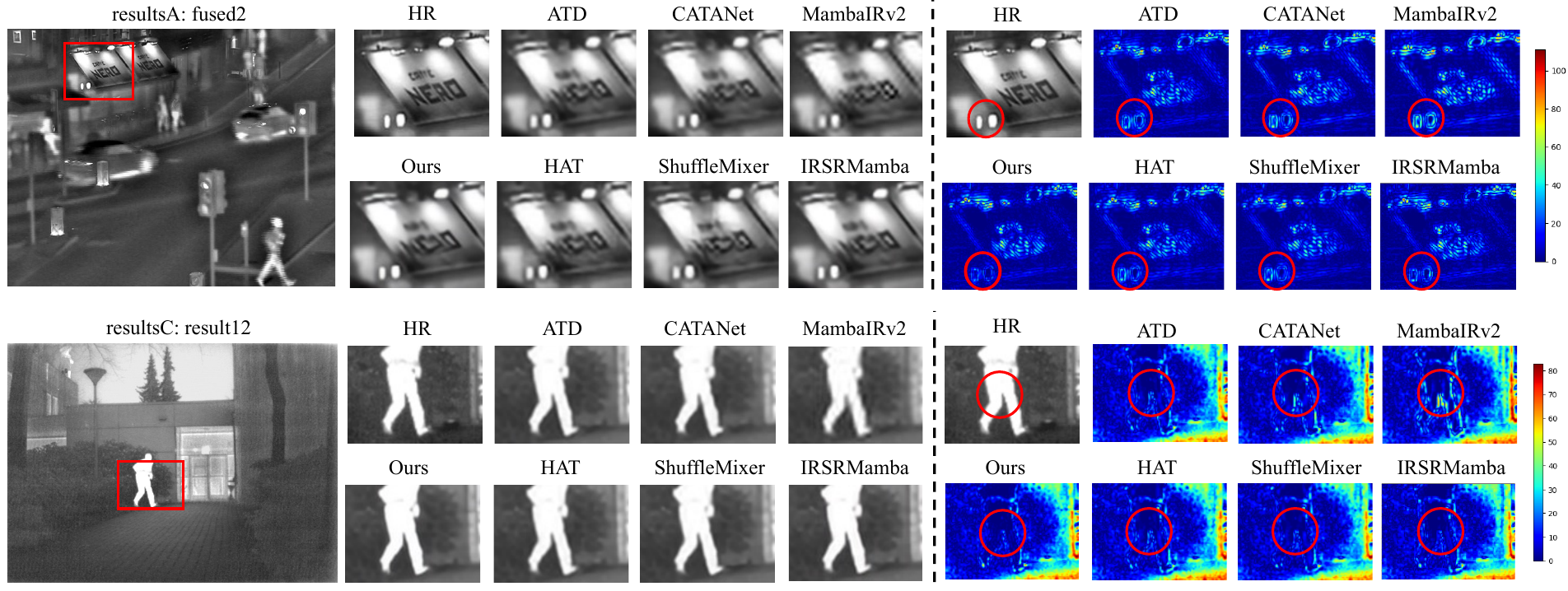}

    \caption{Visual comparison for $\times 4$ super-resolution on images fused2 (top) and result12 (bottom). Ours GPSMamba reconstructs significantly sharper details, a claim substantiated by the minimal reconstruction errors (darker blue) shown in the corresponding residual maps. Best view in zoom.}
    \label{fig:vis} 
\end{figure*}

\begin{table*}[]
\caption{Ablation studies validating our model design. We show the effectiveness of our core components (a) and justify our selection of key hyperparameters (b-d). \textcolor{Gray}{Gray} cells denote the final configuration. Metrics are PSNR/SSIM on result-A/result-C.}
\centering
\resizebox{0.95\textwidth}{!}{%
\begin{tabular}{@{}clllclllclllllll@{}}
\multicolumn{3}{c}{(a)}                                                                                                            &                      & \multicolumn{3}{c}{(b)}                                                                                                           &                      & \multicolumn{3}{c}{(c)}                                                                                                          &                      & \multicolumn{4}{c}{(d)}                                                                                                                                                               \\ \cmidrule(r){1-3} \cmidrule(lr){5-7} \cmidrule(lr){9-11} \cmidrule(l){13-16} 
\multicolumn{1}{l|}{}                            & \multicolumn{1}{c|}{PSNR}               & \multicolumn{1}{c}{SSIM}              & \multicolumn{1}{c}{} & \multicolumn{1}{c|}{Bach size}                  & \multicolumn{1}{c}{PSNR}                & \multicolumn{1}{c}{SSIM}              & \multicolumn{1}{c}{} & \multicolumn{1}{c|}{Blocks}                    & \multicolumn{1}{c|}{PSNR}               & \multicolumn{1}{c}{SSIM}              & \multicolumn{1}{c}{} & \multicolumn{1}{c|}{$\lambda_{\text {freq}}$}                           & \multicolumn{1}{c|}{$\lambda_{\text {phase}}$}                           & \multicolumn{1}{c|}{PSNR}               & \multicolumn{1}{c}{SSIM}              \\ \cmidrule(r){1-3} \cmidrule(lr){5-7} \cmidrule(lr){9-11} \cmidrule(l){13-16} 
\multicolumn{1}{c|}{Mamba (Pure)}                & 39.0346/39.9112                         & 0.9417/0.9528                         &                      & \multicolumn{1}{c|}{16}                         & 38.4182/39.1784                         & 0.9417/0.9440                         &                      & \multicolumn{1}{c|}{4}                         & 38.9902/39.9084                         & 0.9428/0.9536                         &                      & \multicolumn{1}{l|}{0.2}                         & \multicolumn{1}{l|}{0.8}                         & 39.3157/40.2283                         & 0.9440/0.9548                         \\ \cmidrule(r){1-3} \cmidrule(lr){5-7} \cmidrule(lr){9-11} \cmidrule(l){13-16} 
\multicolumn{1}{c|}{$\mathbf{w} \text{ASF-SSM}$}                       & 39.1472/40.0727                         & 0.9429/0.9716                         &                      & \multicolumn{1}{c|}{\cellcolor[HTML]{C0C0C0}32} & \cellcolor[HTML]{C0C0C0}39.1681/40.1465 & \cellcolor[HTML]{C0C0C0}0.9520/0.9547 &                      & \multicolumn{1}{c|}{6}                         & 39.0614/40.0688                         & 0.9433/0.9540                         &                      & \multicolumn{1}{l|}{0.5}                         & \multicolumn{1}{l|}{0.3}                         & 39.3346/40.2236                         & 0.9439/0.9547                         \\ \cmidrule(r){1-3} \cmidrule(lr){5-7} \cmidrule(lr){9-11} \cmidrule(l){13-16} 
\multicolumn{1}{c|}{\cellcolor[HTML]{C0C0C0}$\mathbf{w} \text{ASF-SSM} \& \mathscr{L}_T$} & \cellcolor[HTML]{C0C0C0}39.3142/40.2442 & \cellcolor[HTML]{C0C0C0}0.9440/0.9548 &                      & \multicolumn{1}{c|}{64}                         & 39.0520/40.0754                         & 0.9434/0.9542                         &                      & \multicolumn{1}{c|}{\cellcolor[HTML]{C0C0C0}8} & \cellcolor[HTML]{C0C0C0}39.1030/40.0934 & \cellcolor[HTML]{C0C0C0}0.9442/0.9549 &                      & \multicolumn{1}{l|}{\cellcolor[HTML]{C0C0C0}0.8} & \multicolumn{1}{l|}{\cellcolor[HTML]{C0C0C0}0.2} & \cellcolor[HTML]{C0C0C0}39.3410/40.2216 & \cellcolor[HTML]{C0C0C0}0.9439/0.9547 \\ \cmidrule(r){1-3} \cmidrule(lr){5-7} \cmidrule(lr){9-11} \cmidrule(l){13-16} 
\end{tabular}%
\label{tab:Ablation}
}
\end{table*}

\begin{figure}[t]
\centerline{\includegraphics[width=0.7\columnwidth]{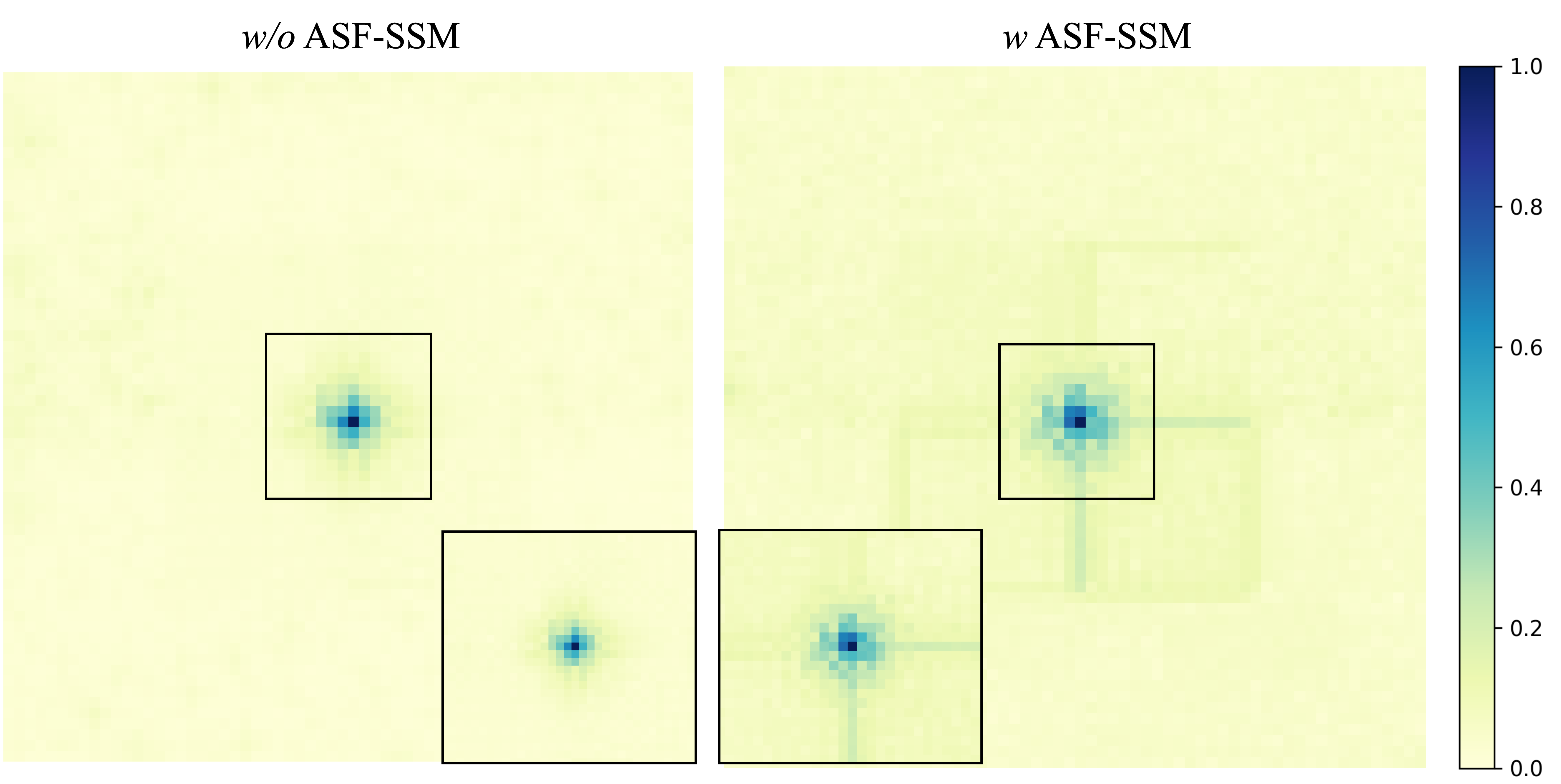}}
\caption{Impact of the Axial Shifted Fusion State Space Module (ASF-SSM) on the Effective Receptive Field (ERF).}
\label{Effective Receptive Field}
\end{figure}

\textbf{Qualitative Results.} As in Fig.\ref{fig:vis}, qualitative comparison for $\times 4$ super-resolution on the result-A/result-C. Our model, GPSMamba, excels in restoring sharp textures and object contours where other methods fail. In the fused2 scene (top), it renders crisp, legible text on the sign, a detail completely lost to blurring in competing results. Similarly, in result12 (bottom), it preserves the pedestrian's sharp silhouette. This superior fidelity is further substantiated by the residual maps (right), where our method exhibits minimal reconstruction error (darker blue) in these critical high-frequency regions, confirming a more accurate reconstruction. 

\textbf{Ablation Experiments.} As shown in Table.\ref{tab:Ablation}(a), we begin with a Mamba (Pure) baseline. Integrating our proposed ASF-SSM yields a significant performance boost, increasing PSNR from 39.03 to 39.14 and SSIM from 0.9528 to 0.9716 on result-A/result-C. This demonstrates the module's efficacy in capturing critical image features. By further incorporating our total loss function ($\mathscr{L}_T$), which includes spectral constraints, the performance is elevated to its peak (39.31/40.24 PSNR), confirming that each component contributes synergistically to the final result. We also analyze the impact of key hyperparameters. As detailed in  Table.\ref{tab:Ablation}(b), we test batch sizes of {16, 32, 64}. The model achieves optimal performance with a batch size of 32, which we adopt for all experiments. Table.\ref{tab:Ablation}(c) shows the effect of varying the number of our core blocks. Performance consistently improves as the model deepens from 4 to 8 blocks, with 8 blocks delivering the best results. Finally, we study the balance between the frequency ($\lambda_{\text{freq}}$) and phase ($\lambda_{\text{phase}}$) loss terms in Table.\ref{tab:Ablation}(d). The results indicate that placing a higher emphasis on the frequency component is beneficial, with the optimal configuration being $\lambda_{\text{freq}}=0.8$ and $\lambda_{\text{phase}}=0.2$. 

\begin{figure}[t]
\centerline{\includegraphics[width=0.7\columnwidth]{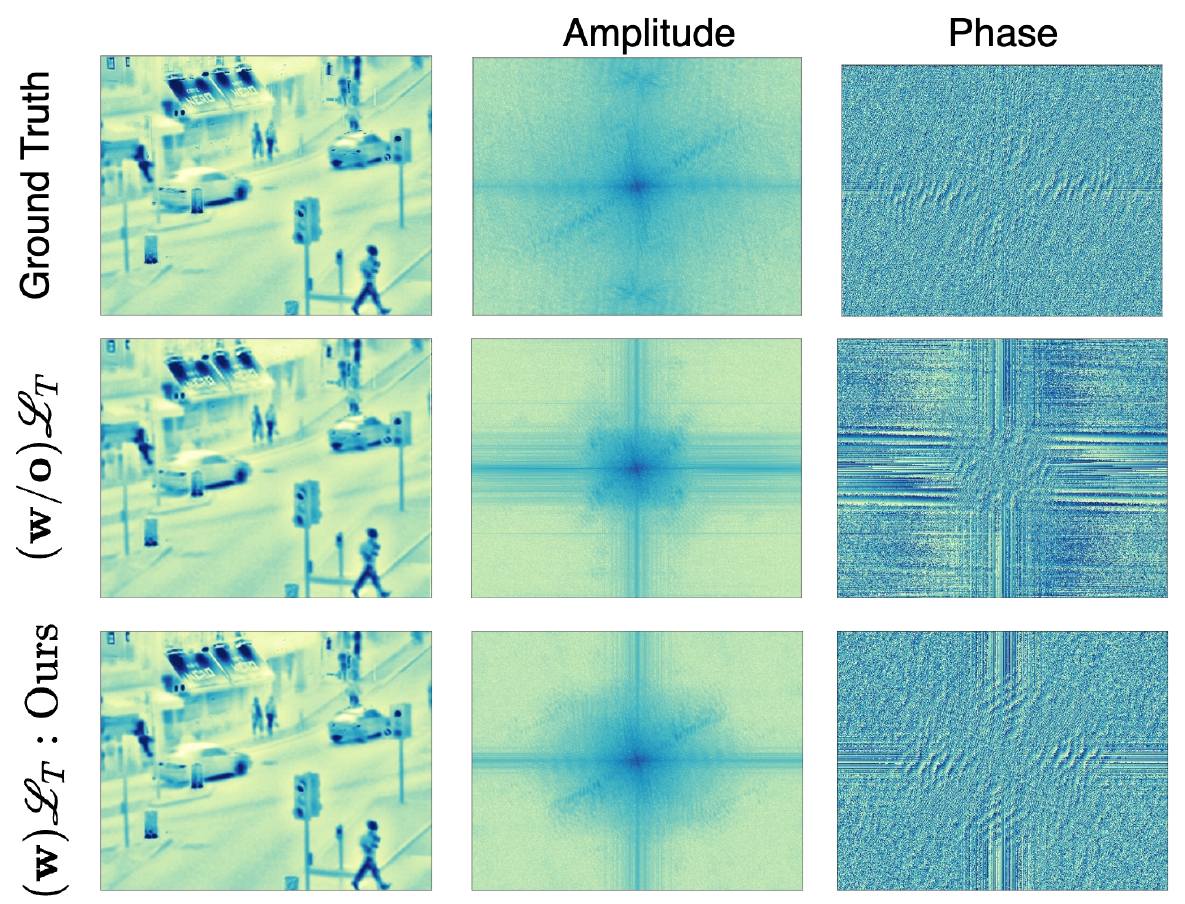}}
\caption{Visual comparison in spatial and frequency domains. From top to bottom: Ground Truth, baseline model (w/o $\mathcal{L}_{\text{T}}$), and our full model (w $\mathcal{L}_{\text{T}}$).}
\label{FFT1}
\end{figure}

\textbf{Effect of ASF-SSM and TSAPC Loss.} We visually validate our core components through an ablation study. As shown in Fig.\ref{Effective Receptive Field}, our ASF-SSM transforms the model's Effective Receptive Field\cite{ding2022scaling,li2023efficient} from a compact, localized region (w/o ASF-SSM) to an expanded and anisotropic one (w ASF-SSM), confirming its capacity for capturing long-range dependencies. Concurrently, the frequency analysis in Fig.\ref{FFT1} demonstrates that our TSAPC loss is critical for structural fidelity. Without it (w/o $\mathcal{L}_{\text{T}}$), the reconstructed phase spectrum exhibits severe artifacts. By enforcing spectral consistency, our method (w $\mathcal{L}_{\text{T}}$) produces a phase spectrum that faithfully aligns with the ground truth, leading to perceptually superior reconstructions.

\section{Conclusion}
\label{sec:Con}
We proposed GPSMamba, an IRSR architecture designed to overcome the global context fragmentation inherent in the 1D causal scanning of SSMs. GPSMamba introduces two core innovations: the ASF-SSM, which injects a fused semantic systematic prompt to guide reconstruction with non-local context, and the TSAPC Loss ($\mathcal{L}_T$), which provides explicit, non-causal supervision to enforce structural and spectral fidelity. Our proposed GPSMamba also achieves state-of-the-art performance, demonstrating that the causal modeling limitations of SSMs can be effectively overcome for visual representation learning.

\bibliography{aaai25}

\end{document}